\documentclass[10pt,twocolumn,letterpaper]{article}
\usepackage[pagenumbers]{cvpr}
\definecolor{cvprblue}{rgb}{0.21,0.49,0.74}
\usepackage[pagebackref,breaklinks,colorlinks,allcolors=cvprblue]{hyperref}
\hypersetup{pdftitle={Learning Capability Boundaries of Vehicle Perception under Adverse Weather},
  pdfauthor={Hangyu Li, Bofeng Cao, Keke Long, Hang Zhou, Shu Han, Zheng Li, Xiaopeng Li}}
\title{Learning Capability Boundaries of Vehicle Perception under Adverse Weather}
\makeatletter
\patchcmd{\@maketitle}{\vspace*{24pt}}{\vspace*{10pt}}{}{}
\makeatother
\author{Hangyu Li\quad Bofeng Cao\quad Keke Long\quad Hang Zhou\quad Shu Han\quad Zheng Li\quad Xiaopeng Li\\{\normalsize University of Wisconsin--Madison}}
\begin{document}
\maketitle
\begin{abstract}
Artificial intelligence (AI) models can generalize to scenarios in which they have never been tested. A system that relies on such a model must determine under which conditions the model still meets its task requirements. Especially, a model can assign high confidence to an incorrect output, and models operating under identical conditions can have different capability boundaries. We try to address this question in vehicle perception, where a perception model must operate under combinations of weather and illumination rarely observed on the same road. Therefore, we introduce ALONSO, a framework for learning where a specified perception model meets a required level of object detection performance. Considering the scarcity of real data and the inability to control weather independently, ALONSO first expands the KITTI dataset with quantitatively controlled weather edits and uses a Gaussian process to actively select which weather settings to query to accelerate the boundary learning. For three-weather combinations, querying 69.3 of 729 settings on average allows ALONSO to learn sufficiently accurate boundaries, accelerating over 10-fold compared to dense queries. To transfer the boundary to a new scene, ALONSO uses DINOv2 to encode scenes and train decoders for capacity boundaries to weather. With references from visually similar reference frames, it correctly predicted perception model capability in 77.76\% of weather conditions without any target-frame tests and in 81.95\% after just four local queries.
\end{abstract}

\begin{figure}[!t]\centering\includegraphics[width=\columnwidth]{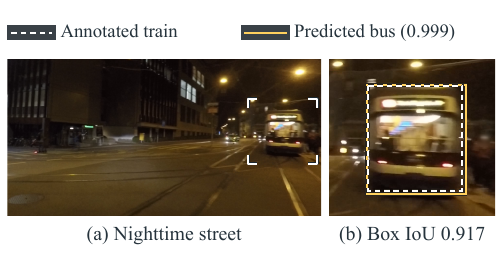}
\caption{Faster R-CNN confidently labels a train as a bus in an ACDC night scene~\cite{sakaridis2021acdc}.}
\label{fig:night_main}\end{figure}
\vspace{-\baselineskip}
\section{Introduction}
When one large language model (LLM) agent builds on another agent's answer~\cite{autogen}, or a robot planner calls a manipulation skill~\cite{saycan}, the next decision depends on what the selected component can do under the conditions at hand. In the safety-critical setting of autonomous driving, a planner relies on perception to locate other vehicles. For a given stretch of road, we seek to predict whether a particular vehicle detector will meet its performance requirement under combinations of weather and illumination it has yet to encounter there. The detector's \emph{capability boundary} separates the conditions in which it meets this requirement from those in which it falls short.

Consider the night scene in Fig.~\ref{fig:night_main}(a). In the enlarged view (b), the Faster R-CNN box overlaps the annotated train at an intersection-over-union of 0.917, yet the detector labels it a bus with confidence 0.999. Even a calibrated confidence score describes the reliability of a box that the detector has emitted~\cite{jiang,kuppers,kuzucu}. A missed vehicle has no box to score, and an untested weather condition has yet to produce any boxes at all. To anticipate performance in that condition, we need evidence about the detector's response as weather changes.

\begin{figure*}[t]
\centering\includegraphics[width=.94\textwidth]{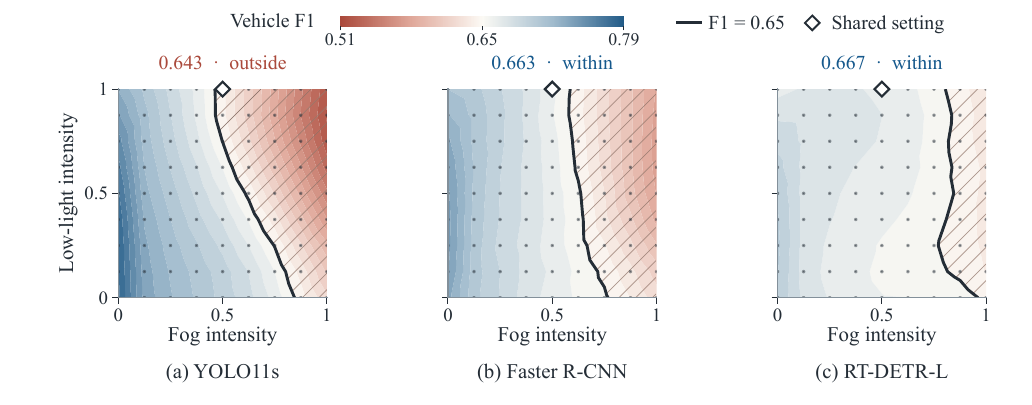}
\caption{Weather capability boundaries for three detectors on the same KITTI scenes. Hatching marks conditions below the common vehicle F1 requirement of 0.65.}
\label{fig:surfaces}
\end{figure*}

Which conditions fall inside the boundary also depends on the detector. At the fog and low-light setting marked by the diamond in Fig.~\ref{fig:surfaces}, two detectors meet a common vehicle F1 requirement and a third falls short on the same labeled roads. Real-world datasets document adverse weather on the road~\cite{sakaridis2021acdc,bijelic2020stf,broedermann2024muses}. Repeated drives show how the same stretch of road looks under different weather and lighting conditions~\cite{cats}, although the vehicles present can change between visits. Controlled edits let us vary fog, rain and illumination independently while keeping the traffic scene fixed. With nine levels for each of three factors, testing every combination requires 729 evaluations per detector. This cost motivates selecting the most informative tests.

\begin{figure*}[t]
\centering\includegraphics[width=\textwidth]{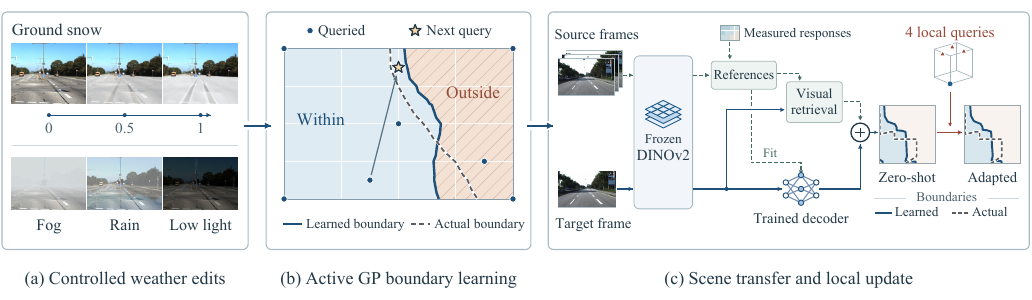}
\input{framework_caption}
\label{fig:framework}
\end{figure*}

ALONSO tests a detector on controlled edits of the same road scene (Fig.~\ref{fig:framework}(a)). It selects the next weather condition according to how much the measurement is expected to resolve uncertain decisions across the grid. In Fig.~\ref{fig:framework}(b), the next query (star) lies where the learned boundary and the dense reference disagree. Once reference views have been measured, ALONSO uses a clear image of a new frame on the same route to predict its boundary, then refines that estimate with four local measurements (Fig.~\ref{fig:framework}(c)).

We first measure how often substituting another detector's boundary changes the decision at a weather setting. We then evaluate how many measurements the specified detector needs to recover its boundary and how reliably it can stop using only the information available during acquisition. Finally, we measure the accuracy of transferring a boundary to a new view on a known route, both before any local test and after four local measurements. The first two evaluations use controlled edits of a fixed KITTI cohort. The third uses frames held out within drives.

\section{Related work}
Prior work measures weather intensities at which detectors first fail, reconstructs detection performance surfaces from sparse samples, and uses earlier drives to predict perception performance. These results leave open how reliably a limited number of tests can identify the combinations of weather and illumination under which a specified detector meets its task requirement on a given set of road scenes. When the view changes, the question is how much of that boundary carries over and how accurately a few local measurements can recover it for the new view.

\subsection{Evaluating perception under adverse weather}
ACDC pairs adverse-condition images with corresponding normal-condition views~\cite{sakaridis2021acdc}. CATS-V2V records ten drives at each of ten road locations under varying weather and lighting conditions, with all visits to a location referenced to the same HD map~\cite{cats}. Rendering lets researchers vary the weather while holding the scene fixed. Foggy Cityscapes transfers labels to rendered fog for training~\cite{sakaridis2018foggy}. Physics-based rain rendering measures how perception changes with rainfall~\cite{halder2019rain,tremblay2021rain}. SCROD uses controlled image generation to expose detector errors across object configurations~\cite{scrod}.

Pettersen and Zhu already compare detectors by finding the first intensity at which detection fails on each image and averaging that intensity across images for each perturbation~\cite{pettersen}. A first-failure intensity for fog and another for rain leave open which mixtures of the two meet a specified vehicle F1. We hold the labeled scenes fixed, vary weather and illumination jointly, and map the combinations at which each frozen detector meets the same absolute F1 requirement. This makes detector specificity testable at each weather setting. Substituting one detector's boundary for another's reveals how often the substitution reverses the decision.

\subsection{Active testing of performance boundaries}
Wang et al. progressively sample brightness and contrast on VOC2007 to reconstruct a detector's performance surface with a multi-kernel GP~\cite{wang}. For a detector subject to a minimum F1 requirement, the location of a reconstruction error matters. A small error near the threshold can reverse the decision, while a larger error far from it can leave the decision unchanged. Active boundary search has direct precedents in driving. HiddenGems maps where simulated maneuvers meet criteria such as collision avoidance and lane keeping~\cite{hg}. GUARD partitions scenario space into safe, unsafe and unresolved regions~\cite{guard}. ODCNet uses active testing and GP residual correction to predict failure from operating conditions, including in a vehicle-detection case study~\cite{odcnet}.

In our setting, each query runs one detector on the same labeled scenes at a chosen weather combination. We adapt stepwise uncertainty reduction for excursion sets~\cite{bect,chevalier} to choose the measurement expected to resolve the most uncertainty about F1 threshold decisions across the weather grid. At an untested combination, the fitted surface predicts whether the detector meets the requirement and how far the nearest threshold crossing lies in weather coordinates. The sampling rule decides when to stop from the measurements available at each step. We then use the dense reference to count missed violations and false alarms at stopping.

\subsection{Predicting perception performance across scenes}
Gur\u{a}u et al. use location and visual similarity to predict perception performance on a repeat traversal from earlier visits~\cite{gurau}. Other methods estimate detection performance on unlabeled target images from prediction consistency~\cite{yoo} or box stability under feature dropout~\cite{yangdrop}. These approaches estimate how a detector will perform on a visit or an image set. For a clear view on a known drive, we seek a prediction that varies with the weather we might apply to that same view, before the detector sees any of those edits.

ALONSO uses the clear target frame's DINOv2 representation~\cite{oquab2023dinov2}, the detector identity and reference frames from the same drive to predict F1 across combinations of fog, low light and rain. This transfers a weather response to the new view and gives an initial boundary without a target measurement. Four local measurements then reveal where the prediction differs from the observed F1. Relationships between prediction errors learned on reference frames carry those corrections to weather settings held out from local testing.

\section{Learning a queryable capability boundary}
Given a frozen detector, a set of labeled road images and a performance requirement, ALONSO estimates which weather settings meet the requirement. We first define what a test measures, then describe how to choose tests and reuse measurements when the target changes from a cohort to an individual frame.
\subsection{Weather edits and detector measurements}
We cache monocular depth and semantic masks for each KITTI frame. Fog, low light, rain, falling snow and ground snow each take nine intensities from 0 to 1. In linear RGB we edit ground snow and wetness, then illumination, atmospheric attenuation and visible precipitation. For fog, transmittance at depth $d$ and intensity $s$ is $\exp(-0.06sd)$. A fixed per-frame seed generates a reusable particle pool, so increasing precipitation activates particles on the same scene.

Fix a detector $d$, a labeled scene cohort $S$, and a family of weather edits indexed by $w\in[0,1]^k$, with $k\in\{1,2,3\}$. At a selected coordinate we apply the corresponding joint edit to every frame in $S$, retaining its original vehicle annotations. We run the frozen detector at score threshold 0.25 and match vehicle detections to ground truth at intersection-over-union 0.50. We pool counts over the cohort before computing
\begin{equation}
 f_{d,S}(w)=\frac{2\,\mathrm{TP}}{2\,\mathrm{TP}+\mathrm{FP}+\mathrm{FN}} .
\end{equation}
An empty ground-truth set with false positives scores zero; an all-zero denominator is undefined and excluded. For KITTI we group Car, Van and Truck ground truth, retain boxes at least 25 pixels tall, and group predicted car, bus and truck classes. A query reveals one value of $f$ for one detector at one coordinate, at the cost of evaluating the cohort there.

For a required F1 of $t$, the acceptable set is $\{w:f_{d,S}(w)\geq t\}$. We use $t=0.65$, the highest multiple of 0.05 reached by all three detectors on the clean KITTI cohort. We use Ultralytics checkpoints for YOLO11s~\cite{yolo11release} and RT-DETR-L~\cite{zhao2024rtdetr}. For Faster R-CNN~\cite{ren2015faster}, we use the TorchVision ResNet-50-FPN v2 model with its COCO weights. Clean F1 is 0.782 for YOLO11s, 0.734 for Faster R-CNN and 0.699 for RT-DETR-L. This common working point permits the same performance question to be asked of all detectors, including on a drive whose clean score already falls short. A score exactly at the requirement qualifies.

\subsection{Boundary decisions, performance margins, and distances}
A Gaussian process (GP) with a squared-exponential kernel on normalized coordinates supplies estimates at unmeasured grid nodes. Nearby edits on the same scenes often produce similar F1, so a measurement can inform predictions at neighboring settings. The posterior uncertainty also supplies the acquisition rule below. All acquisition methods share the kernel length scale 0.24 and observation noise parameter 0.035.

We retain exact measurements at queried nodes and GP means at the others. We divide each grid cell into simplices using a consistent coordinate ordering and interpolate affinely within each simplex. Denote this continuous surface by $\hat f$ and its in-domain level set by $B_t=\{u:\hat f(u)=t\}$. The same surface supplies the decision $\hat f(w)\geq t$, the performance margin $h(w)=\hat f(w)-t$, and the signed distance
\begin{equation}
 D(w)=\operatorname{sgn}(h(w))\,\min_{u\in B_t}\|w-u\|_2 .
\end{equation}
The intersection with a simplex is a point, line segment or planar facet according to dimension. Minimizing over their union gives the closest in-domain interface. An entirely qualifying or entirely failing domain returns that state and an undefined distance. A margin of $h=0.03$ places predicted F1 three points above the requirement. The distance magnitude gives the smallest change in normalized weather coordinates needed to reach the estimated boundary. A small margin can correspond to a large distance where the response varies slowly.

\subsection{Active query selection and stopping}
A useful observation resolves uncertainty about which side of the requirement other weather conditions occupy. Following stepwise uncertainty reduction (SUR) for excursion sets~\cite{bect,chevalier}, ALONSO chooses the candidate giving the largest expected decrease in the sum of Bernoulli variances of the threshold indicators. For unqueried node $z$, let $p_z=\Phi(a_z)$, where $a_z$ is the posterior mean minus $t$, divided by the posterior standard deviation scaled using observed residuals. With $U=\sum_z p_z(1-p_z)$ we choose
\begin{equation}
 x^*=\arg\max_{x\in C}\big[U-\mathbb{E}_{y_x}U_{\rm updated}(x,y_x)\big].
\end{equation}
The candidate list combines the eight nodes nearest the threshold in standardized units and the eight with largest posterior standard deviation. For each candidate we hold the current trend and scale fixed and use four-point quadrature to evaluate the expected reduction across all unqueried nodes. We then measure the selected node and refit.

A stopping decision must be available before the remaining weather settings are measured. After initialization and each subsequent update, we count the unmeasured nodes whose 90\% posterior intervals contain $t$. Three consecutive counts at or below 10\% of the entire grid trigger a stop. Otherwise sampling continues to the query limit (the maximum number of queries per run). We evaluate the decisions at stopping against the withheld measurements.

\subsection{Scene transfer and local updates}
Let $I$ be a clear image of a held-out frame and let $f_{d,I}(w)$ be the detector's F1 on that frame after applying weather $w$. Reference frames from the same collection supply both clear images and measured fog--low-light--rain response fields. A frozen DINOv2 ViT-S/14 backbone from the Depth Anything V2 Metric Outdoor Small checkpoint~\cite{oquab2023dinov2,depthanythingv2} encodes each clear frame. In each split we fit a 32-dimensional feature transform on training frames, then train a single decoder shared across the drives. Its inputs are the transformed embedding and detector identity. Two hidden layers (96 and 64 units) combine the feature with a three-way detector identifier. The output layer is initialized at the detector's training mean field and trained against measured F1 on reference frames. Its 729 outputs estimate F1 at the weather grid nodes.

To adapt this field to the target view, we retrieve visually similar reference frames from its drive in the whitened feature space. The circled plus in Fig.~\ref{fig:framework}(c) combines the decoded field with either the neighbors' measured F1 values or their measured residuals relative to the decoder. Neighbor weights are proportional to inverse squared feature distance. The neighbor count, mixing weight and combination rule are selected on source validation frames for each detector and fold. Thresholding the resulting field at 0.65 predicts which weather conditions meet the requirement.

Four target queries measure the clear condition and three joint weather settings. Write $p$ for the transferred field and $Q$ for the queried coordinates with defined target F1. The discrepancies $f_{d,I}(Q)-p(Q)$ reveal how the target differs from its references. From training-frame residuals we estimate their second moment across weather coordinates for each detector. Denote the regularized matrix by $C$. For unmeasured coordinates $A$ the update is
\begin{equation}
\hat f_{d,I}(A)=p(A)+C_{AQ}C_{QQ}^{-1}\big(f_{d,I}(Q)-p(Q)\big).
\end{equation}
Thus a measured discrepancy carries more weight at weather settings with related source residuals. Source validation selected 2\% diagonal shrinkage in every fold and positive budget. Predictions are clipped to $[0,1]$. An undefined F1 costs one query and contributes no residual. At zero queries the target contributes its clear image and drive identity, and the detector is specified. Its clear-condition F1 first becomes available at one query.

\section{Experiments}
\begin{figure*}[t]\centering
\includegraphics[width=\textwidth]{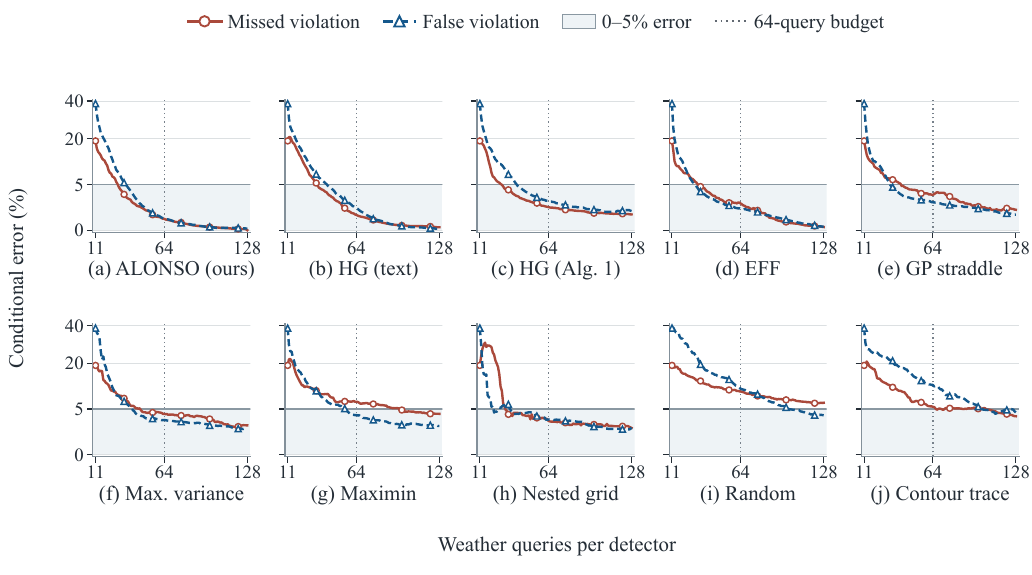}
\caption{Missed violations (circles) and false alarms (triangles) as queries accumulate on three-factor weather grids.}
\label{fig:policycurves}\end{figure*}

We use all 7,481 labeled KITTI training images~\cite{kitti} to evaluate three frozen detectors on the same scenes. A weather family specifies which edits can vary. The five factors are fog (F), low light (L), rain (R), snowfall (S) and ground snow (G). The design uses one falling precipitation type per scene. Ground snow describes snow already on the road, so rain can fall on it. This rule excludes the rain--snowfall pair and the three triples that contain it. The remaining nine pairs and seven triples are listed below.

\begin{center}\small
\begin{tabular}{@{}lll@{}}\toprule
Falling precipitation & Two factors & Three factors \\
\midrule
None & FL, FG, LG & FLG \\
Rain & FR, LR, RG & FLR, FRG, LRG \\
Snowfall & FS, LS, SG & FLS, FSG, LSG \\
\bottomrule\end{tabular}
\end{center}

Each active factor takes nine levels from 0 to 1 in steps of $1/8$. Inactive factors stay at zero. A grid face at zero is shared with a lower-dimensional family. Counting each setting once gives $1+5(8)+9(8^2)+7(8^3)=4{,}201$, including the clear condition. Across the three detectors and 21 families, 43 of the 63 response grids cross the F1 requirement of 0.65 (5 one-factor, 19 two-factor and 19 three-factor cases). The other 20 meet the requirement throughout. Both directional error rates can be measured on the 43 crossing grids.

We compare acquisition policies by revealing entries of the measured grid one at a time. Within each paired trial all policies start at the grid center and corners, plus one random remaining node in one dimension or two in higher dimensions. These 4, 7 and 11 initial measurements count toward the fixed budgets of 8, 24 and 64. The respective query limits are 9, 48 and 128. Each policy sees F1 only at its chosen nodes. We average over ten initializations within each detector--family pair, then weight pairs equally. To score predictions between grid nodes, we use 304 previously examined development coordinates rendered on the same scenes with the same seeds. These yield 912 distinct detector--weather answers and 9,120 predictions across initializations.

\begin{table*}[t]\centering\small
\caption{Queries and boundary decisions for all ten acquisition policies. The 1D--3D columns give the mean number of queries and, in parentheses, the number of runs reaching the retrospective target. Runs that do not reach it count at the corresponding query limit. Grid error averages crossing cases at fixed budgets. Off-grid errors use 9,120 predictions. Bold and underlining mark the best and next distinct values.}
\label{tab:acquisition}\begingroup\setlength{\tabcolsep}{4pt}
\begin{tabular*}{\textwidth}{@{\extracolsep{\fill}}lrrrrr@{}}\toprule
 & \multicolumn{3}{c}{Mean queries $\downarrow$ / runs reaching target $\uparrow$} & \multicolumn{2}{c}{Side error (\%) $\downarrow$} \\
\cmidrule(lr){2-4}\cmidrule(lr){5-6}
Acquisition & 1D (50) & 2D (190) & 3D (190) & Grid & Off-grid \\
\midrule
\textbf{ALONSO (ours)} & \underline{5.46} (\textbf{50}) & \textbf{17.84} (\textbf{190}) & \textbf{67.50} (\textbf{190}) & \underline{0.148} & \textbf{0.252} \\
HiddenGems (text)~\cite{hg} & \textbf{5.42} (\textbf{50}) & 20.79 (\textbf{190}) & 76.17 (\textbf{190}) & \textbf{0.141} & \underline{0.285} \\
HiddenGems (Alg.~1)~\cite{hg} & 7.02 (\textbf{50}) & 21.88 (\underline{187}) & 92.90 (160) & 1.394 & 0.746 \\
EFF~\cite{eff} & 5.54 (\textbf{50}) & \underline{19.06} (\textbf{190}) & \underline{68.87} (\textbf{190}) & 0.514 & 0.471 \\
GP straddle~\cite{bryan} & 5.74 (\textbf{50}) & 22.76 (\textbf{190}) & 87.95 (\underline{174}) & 1.253 & 0.384 \\
Maximum variance~\cite{krause2008entropy} & 6.52 (\textbf{50}) & 27.79 (178) & 113.17 (120) & 1.926 & 0.844 \\
Maximin distance~\cite{gonzalez1985farthest} & 7.92 (\textbf{50}) & 30.28 (167) & 117.33 (116) & 1.856 & 1.129 \\
Nested grid & 7.60 (\textbf{50}) & 28.87 (181) & 108.25 (150) & 1.912 & 0.954 \\
Random & 6.82 (\textbf{50}) & 32.68 (174) & 124.02 (58) & 2.701 & 1.447 \\
Contour tracing & 5.58 (\textbf{50}) & 29.81 (186) & 127.89 (1) & 3.145 & 2.050 \\
\bottomrule\end{tabular*}\endgroup

\end{table*}
\subsection{Detector-specific weather boundaries}

At fog intensity 0.5 and low-light intensity 1.0, YOLO11s achieves pooled F1 of 0.643, Faster R-CNN 0.663, and RT-DETR-L 0.667. The detector with the highest clean score is the only one below the requirement (Fig.~\ref{fig:surfaces}). Across all 4,201 unique conditions, the corresponding detectors meet the common requirement at 3,007, 2,266 and 489 coordinates.

To test dependence on the detector across coordinates, we freeze the nested sampling schedule and exchange fitted surfaces. For each target we evaluate both alternatives as donors. We also pool observations to estimate a shared shape with a target-specific offset. At a total budget of $q$, each donor supplies $\lfloor q/3\rfloor$ measurements and the target supplies the remainder; a more expensive pooled comparison uses $q$ measurements per detector. All comparisons answer the same off-grid questions. Table~\ref{tab:swap} separates the cost of reusing a surface from that of estimating its target-specific level.

\begin{table}[htb]\centering\small
\caption{Off-grid decisions after changing which detector supplies the response. Cost counts detector--condition evaluations. Each target has two donors in the swap test.}
\label{tab:swap}
\begin{tabular}{@{}lrrr@{}}\toprule
Surface & Cost & Wrong / decisions & Error\\\midrule
Target detector & $q$ & 87 / 9,120 & \textbf{0.95}\%\\
Donor detector & $q$ & 6,292 / 18,240 & 34.50\%\\
Shared + offset & $q$ & 1,124 / 9,120 & 12.32\%\\
Shared + offset & $3q$ & 476 / 9,120 & \underline{5.22}\%\\\bottomrule
\end{tabular}
\end{table}
With the sampling schedule held fixed, substituting a donor raises the off-grid error from 0.95\% to 34.50\%. Giving the shared surface a target-specific offset reduces the error to 12.32\% at the matched budget. Even with three times as many measurements it reaches 5.22\%. The same offset applies at every weather coordinate, so it can account for a detector that scores consistently lower but cannot represent a change in the relative response to fog and low light.

\begin{table*}[!t]\centering\small
\caption{Stopping in three-factor weather families with a measured crossing. Each policy has the same 190 trials. Runs that do not trigger count at the 128-query limit. The remaining columns count runs that trigger, meet both directional error targets at stopping, and sustain the recovery criteria through the query limit.}
\label{tab:runtime}\begin{tabular*}{\textwidth}{@{\extracolsep{\fill}}lrrrr@{}}\toprule
Acquisition & Queries $\downarrow$ & Triggered $\uparrow$ & Passed at stop $\uparrow$ & Sustained $\uparrow$ \\ \midrule
\textbf{ALONSO (ours)} & \underline{69.28} & \textbf{190} & \textbf{188} & \textbf{183} \\
EFF~\cite{eff} & \textbf{68.15} & \textbf{190} & 174 & 169 \\
HiddenGems (text)~\cite{hg} & 77.64 & \textbf{190} & \underline{181} & \underline{178} \\
HiddenGems (Alg. 1)~\cite{hg} & 91.04 & 167 & 153 & 141 \\
GP straddle~\cite{bryan} & 82.19 & \underline{181} & 154 & 148 \\
Maximum variance~\cite{krause2008entropy} & 104.08 & 144 & 115 & 102 \\
Maximin~\cite{gonzalez1985farthest} & 106.00 & 139 & 99 & 88 \\
Nested grid & 106.01 & 165 & 160 & 137 \\
Random & 111.74 & 107 & 44 & 42 \\
Contour tracing & 122.49 & 20 & 4 & 0 \\
\bottomrule\end{tabular*}

\end{table*}
\begin{figure}[!t]\centering\includegraphics[width=\columnwidth]{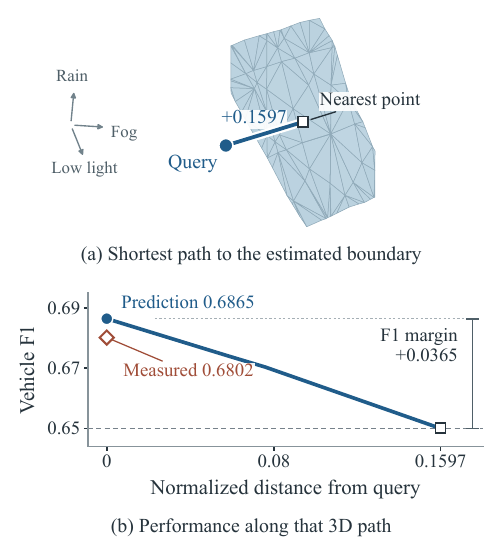}
\caption{For one unmeasured condition, (a) locates the nearest point on the learned boundary and (b) shows the F1 change along that path.}
\label{fig:offgrid_main}\end{figure}
\begin{figure*}[!t]\centering\includegraphics[width=.72\textwidth]{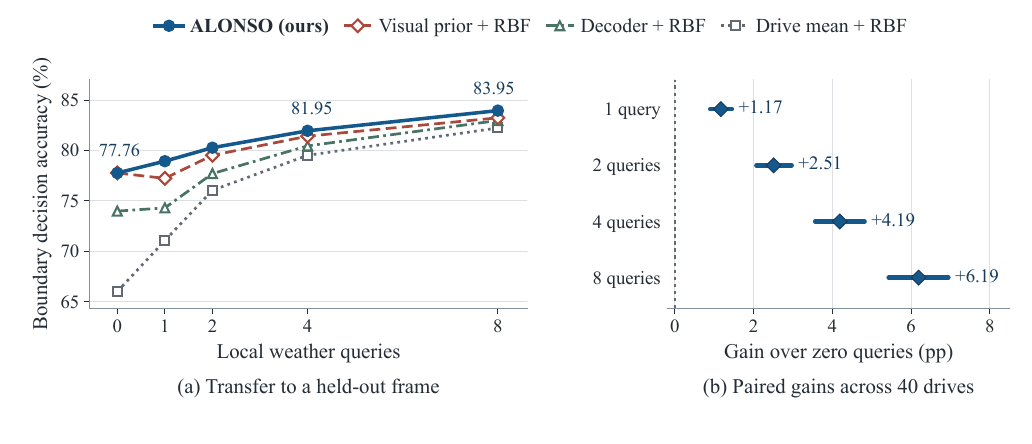}
\caption{On held-out frames of known drives, (a) compares accuracy across query budgets and (b) shows the gain over zero queries with 95\% paired drive-bootstrap intervals.}
\label{fig:frames}
\par\medskip
\includegraphics[width=.72\textwidth]{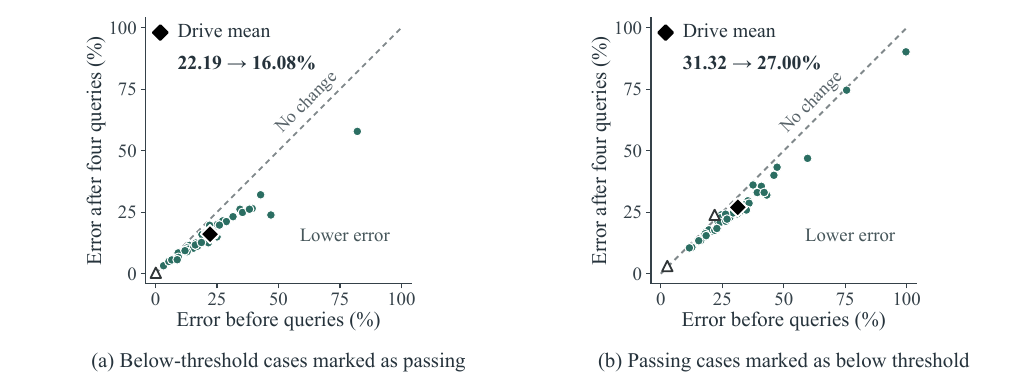}
\caption{Drive-level missed violations (a) and false alarms (b) before and after four local queries. Points below the diagonal indicate lower error after the update.}
\label{fig:transferdirections}\end{figure*}

\subsection{Query efficiency and stopping reliability}
All ten policies use the same GP to reconstruct F1 from their chosen measurements (Table~\ref{tab:acquisition}). Random sampling draws unused settings without replacement. Nested grid refines a coarse lattice, while maximin chooses the setting farthest from its nearest observation~\cite{gonzalez1985farthest}. Maximum variance chooses the largest posterior variance~\cite{krause2008entropy}. These rules distribute tests without using the F1 threshold. Contour tracing first brackets a crossing, then follows neighboring grid cells whose measured corners straddle the threshold.

GP straddle favors conditions whose predicted F1 lies near the requirement relative to their uncertainty~\cite{bryan}. Expected feasibility (EFF) scores the expected proximity of F1 to the threshold within an uncertainty-dependent band~\cite{eff}. HiddenGems alternates between sampling near the predicted threshold and sampling far from previous observations~\cite{hg}. Its prose increasingly favors the threshold as the budget is spent, while the inequality in Algorithm~1 gives the opposite schedule. We evaluate both with the same random draws. \textbf{ALONSO (ours)} scores the expected effect of one measurement on threshold uncertainty at all unmeasured nodes~\cite{bect,chevalier}. A candidate can therefore be valuable because of the other weather combinations it resolves.

\subsubsection{Boundary accuracy at fixed query budgets}
Figure~\ref{fig:policycurves} tracks failures predicted to pass and passes predicted to fail, each as a fraction of its true class, on the 19 three-factor grids with a crossing. Square-root spacing expands the shaded 0--5\% target range, making it easier to compare the two rates at the marked budget of 64 queries. At fixed budgets, the ALONSO boundary gives 23 incorrect decisions on 9,120 off-grid predictions. The same-budget nested grid gives 87 and random sampling gives 132. The two lowest errors on the native grid are 0.141\% for HiddenGems (text) and 0.148\% for ALONSO. Table~\ref{tab:acquisition} reports the fixed-budget errors for all ten policies.

After 64 queries, ALONSO predicts that YOLO11s meets the requirement when fog, low light and rain each have intensity $1/3$. The shortest path from that unmeasured condition to the estimated boundary is 0.1597 normalized weather units (Fig.~\ref{fig:offgrid_main}(a)). Along this path, predicted F1 falls from 0.6865 to 0.65, giving a performance margin of $+0.0365$ at the query (Fig.~\ref{fig:offgrid_main}(b)). A held-out detector evaluation gives F1 of 0.6802, confirming the predicted side.

On the measured off-grid development coordinates, ALONSO incorrectly qualifies 13 of 4,553 predicted passes. Those 13 are also 13 of 4,570 actual failures. Across the 6,888 predictions for which all seven policies in the signed-distance comparison and the dense reference have a finite boundary, ALONSO has a median absolute signed-distance error of 0.00234 in normalized weather coordinates. The reference interpolates the measured grid.

\subsubsection{Recovery cost and online stopping}
For a common quality target, we require at most 5\% wrong decisions on each true side and at most 10\% of grid nodes with unresolved posterior intervals. We record the first measurement after which all three conditions hold through the query limit. This retrospective comparison uses labels from the dense grid and the future trajectory. Across 190 three-factor runs, ALONSO, expected feasibility and HiddenGems (text) all recover sustained quality, after means of 67.50, 68.87 and 76.17 measurements, respectively.

At this retrospective quality criterion, dense grid size divided by mean ALONSO query count is 1.65, 4.54 and 10.80 for one, two and three factors, respectively.

During acquisition, each policy applies the stopping rule using only the measurements available at that step (Table~\ref{tab:runtime}). The query count ends at the trigger or, for runs that never trigger, at the query limit. On three-factor crossing cases, ALONSO triggers in every trial, averaging 69.28 measurements out of 729. At stopping, 188 of 190 trials meet both directional error targets; 183 continue to meet all criteria through the query limit. In the two immediate failures, Faster R-CNN misses one violating setting out of six or thirteen. Small failure regions explain how a modest count of ambiguous settings can still leave a consequential error. EFF triggers earlier on average (68.15 queries) and meets both directional targets in 174 trials.

\subsection{Boundary transfer to held-out frames on known drives}
Five within-drive splits hold out entire frames and their weather measurements, yielding 6,676 targets across 40 KITTI drives. A held-out frame contributes no weather labels to training. At zero queries the estimator uses the clear image, detector identity and the drive's measured references. Four queries reveal F1 at the clear condition and three adverse settings. We score all budgets on the same 176 hidden weather settings and weight each drive equally. The drive mean assigns every target on a drive the same source response at each weather setting. The decoder conditions its estimate on the clear target image. Visual transfer also retrieves measured responses from similar reference frames on that drive. Using the same visual prior with two update rules isolates the contribution of learning how prediction errors covary across weather conditions.

Figure~\ref{fig:frames}(a) compares the four transfer methods on the same held-out frames. Before any target measurement, visual transfer correctly classifies 77.76\% of hidden conditions, compared with 73.97\% for the image decoder alone and 66.00\% for the drive mean. With four local measurements, the source-learned update reaches 81.95\% accuracy, compared with 81.41\% for the fixed radial basis function (RBF) update using the same visual prior. All 40 drives improve over zero queries with the source-learned update. The mean gain is 4.19 percentage points (95\% paired drive-bootstrap interval 3.56--4.82, Fig.~\ref{fig:frames}(b)).

Figure~\ref{fig:transferdirections} shows how this gain changes the two kinds of decision error. Among true failures, the fraction incorrectly passed falls from 22.19\% to 16.08\%, improving on 39 of 40 drives. Among true passes, the fraction incorrectly rejected falls from 31.32\% to 27.00\%, improving on 38 drives. Each rate pools frames and all three detectors within a drive. The four-query estimates average over three query schedules.

\section{Discussion and conclusion}
A weather boundary belongs to a detector, a scene population and a task requirement. On identical images, substituting another detector's boundary can reverse the decision. Measuring the target detector at one setting can resolve uncertainty at neighboring settings. The errors at stopping show that even a small region of failure can matter when few weather settings violate the requirement.

A clear image helps identify reference views for a new frame on the same drive. Local measurements reveal where the target departs from those references. Four local measurements reduce both missed violations and false alarms on average (Fig.~\ref{fig:transferdirections}). Whether these boundaries carry over to naturally occurring weather on a new route remains open.

\clearpage
\section*{LLM Usage Statement}
The authors used OpenAI ChatGPT and Codex to discuss research questions and experimental designs, identify and organize related work, draft and revise manuscript text, and implement and run multi-step tasks for evaluation, analysis and figure preparation. OpenAI image generation was used to explore layouts for the framework illustration. The final figure was redrawn using measured data and source images. The authors are responsible for the experimental evidence, interpretation and final manuscript.
{\small\bibliographystyle{ieeenat_fullname}\bibliography{references}}
\backrefsetup{disable}
\onecolumn
\setcounter{figure}{0}\renewcommand{\thefigure}{S\arabic{figure}}
\setcounter{table}{0}\renewcommand{\thetable}{S\arabic{table}}
\renewcommand{\theHfigure}{supp.fig.\arabic{figure}}
\renewcommand{\theHtable}{supp.tab.\arabic{table}}

\providecommand{\mainThreeFactorFigure}{Fig.~\ref{fig:policycurves}}
\providecommand{\mainCostTables}{Tables~\ref{tab:acquisition} and~\ref{tab:runtime}}
\begingroup
\renewcommand{\twocolumn}[1][]{\noindent\begin{minipage}{\textwidth}#1\end{minipage}\par}
\maketitlesupplementary
\endgroup
\section*{A. Weather controls and the performance requirement}

\subsection*{A.1. From single controls to joint weather}

Figure~\ref{fig:weather_appearance} shows five controls at levels 0, 0.5 and 1 of the nine-level experimental grid. Each KITTI scene keeps its geometry, annotations and particle seed. At the intersection, fog obscures the distant tree line before the near signal pole. On the residential street, snowfall adds particles in front of parked cars and ground snow brightens the road between them.

Rendered rain has been assessed through perceptual judgments~\cite{halder2019rain} and synthetic fog through downstream performance on real roads~\cite{sakaridis2018foggy}. The two ACDC photographs~\cite{sakaridis2021acdc} illustrate loss of distant contrast in real fog and dark wheel paths on snow.

\begin{center}\begin{minipage}{\linewidth}\centering
\includegraphics[width=1\linewidth]{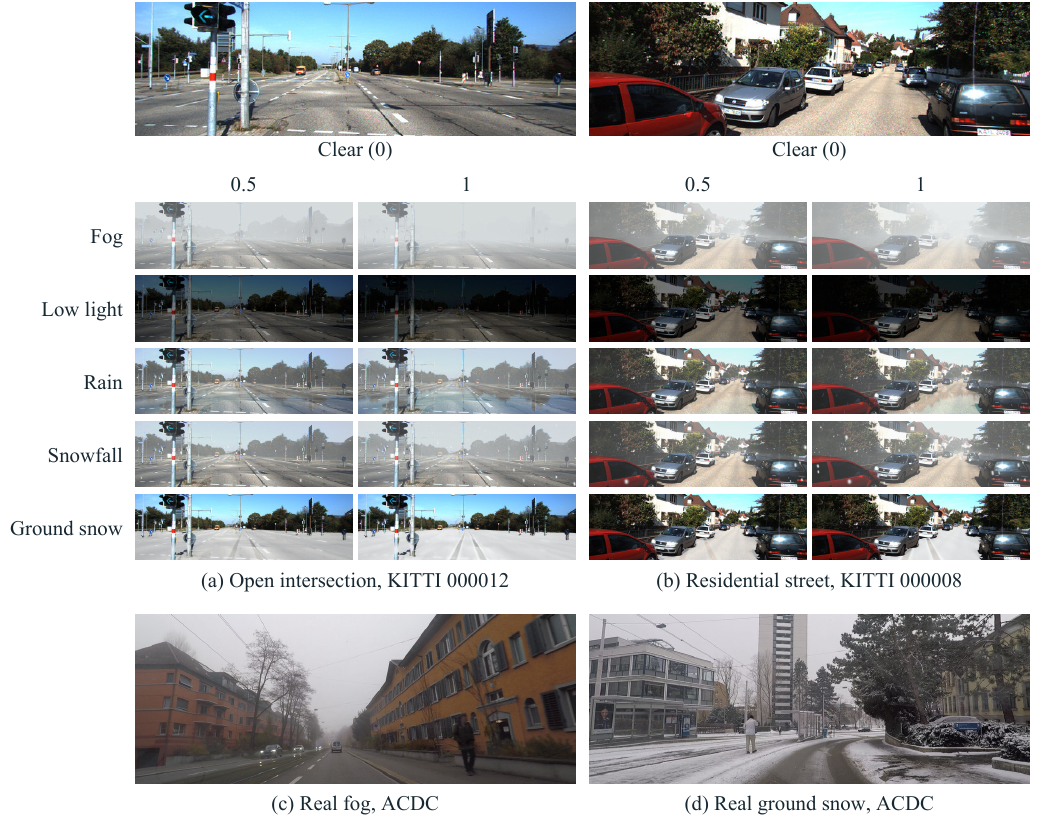}\par
\captionof{figure}{Five controls at two positive intensities beneath the clear KITTI views (a, b). Real ACDC photographs show fog (c) and ground snow (d).}\label{fig:weather_appearance}\end{minipage}\end{center}

Figures~\ref{fig:pairs_intersection} and~\ref{fig:pairs_street} use identical rows and columns for both scenes. In the fog--ground snow row, moving from $(0.5,0.5)$ to $(1,0.5)$ increases fog at fixed ground snow.

\clearpage
\noindent\begin{minipage}{\linewidth}\centering
\includegraphics[width=.99\linewidth]{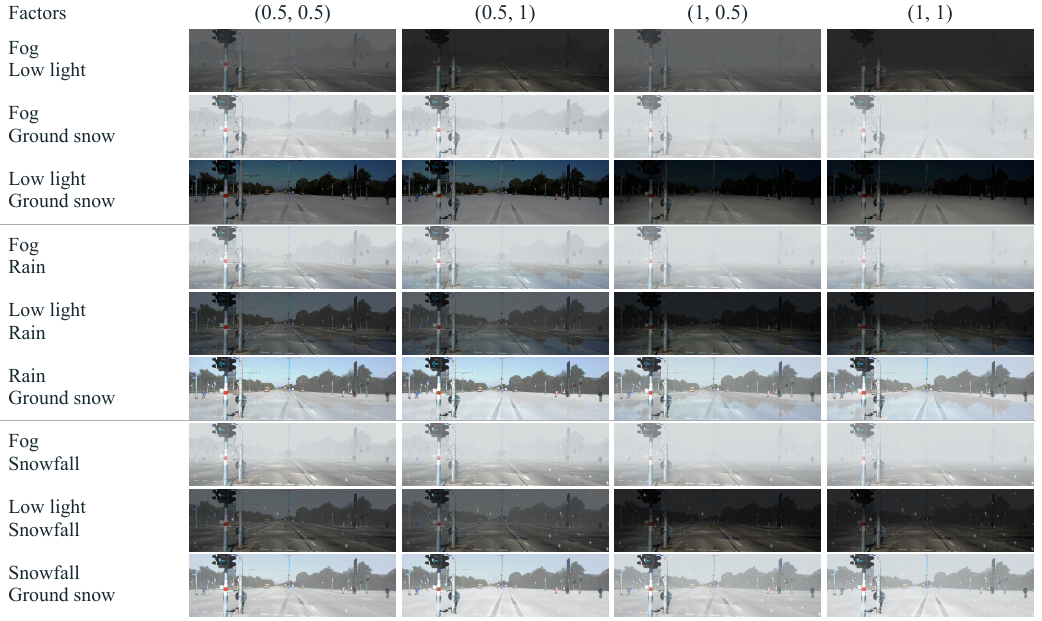}\par
\captionof{figure}{Nine weather pairs on the open intersection (KITTI 000012). Coordinates follow the listed factor order.}\label{fig:pairs_intersection}
\end{minipage}\par\vspace{3pt}

\noindent\begin{minipage}{\linewidth}\centering
\includegraphics[width=.99\linewidth]{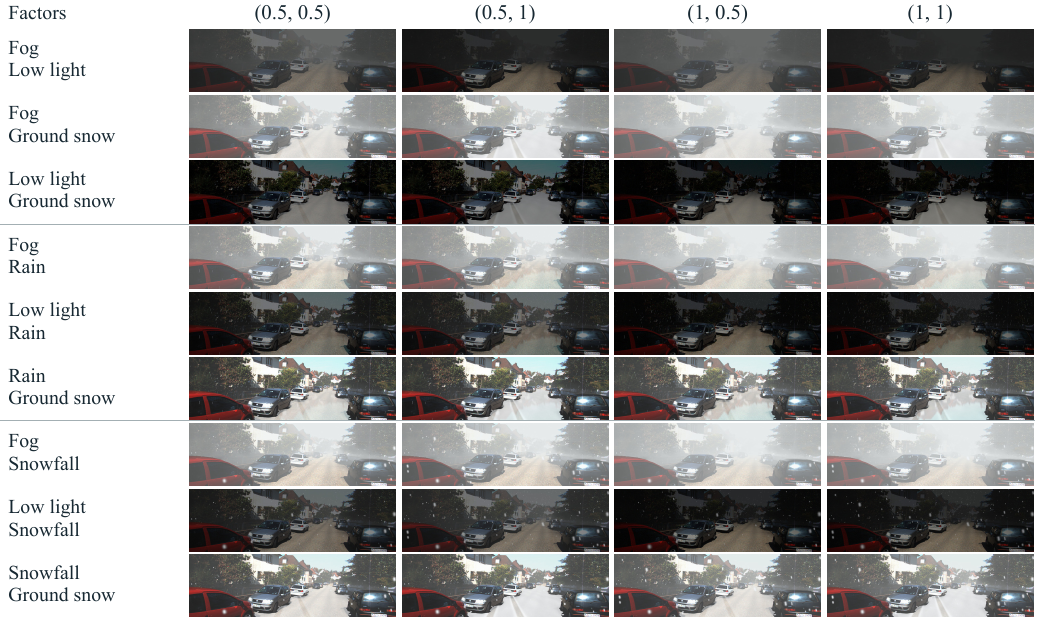}\par
\captionof{figure}{The same nine weather pairs and four intensity settings on the residential street (KITTI 000008).}\label{fig:pairs_street}
\end{minipage}\par

\clearpage
Figures~\ref{fig:triples_intersection} and~\ref{fig:triples_street} introduce a third control. Within each group, the upper and lower rows set the first factor to 0.5 and 1. Columns vary the second and third factors independently. In the fog--low light--rain group, for example, the upper right image has fog 0.5, low light 1 and rain 1.

\begin{center}\begin{minipage}{\linewidth}\centering
\includegraphics[width=1\linewidth]{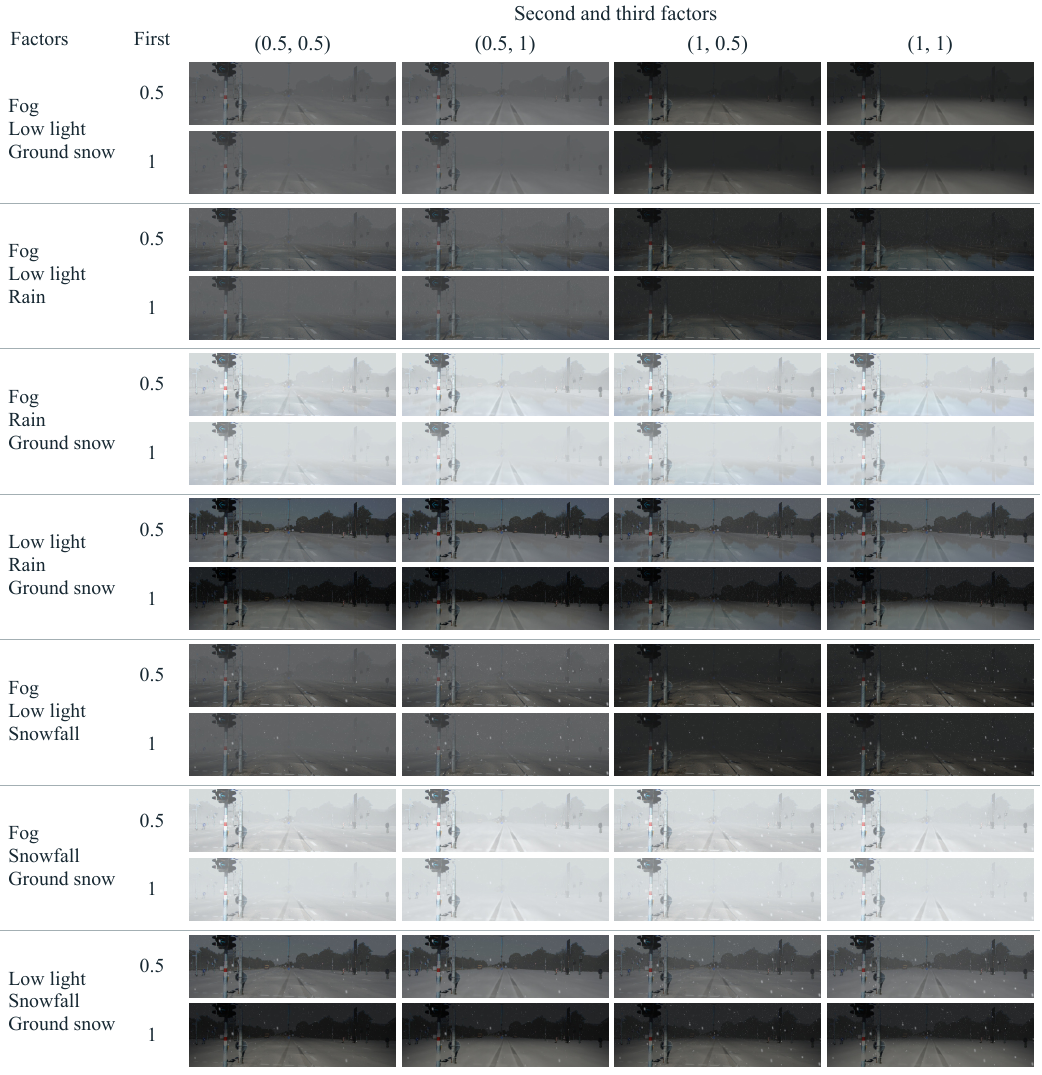}\par
\captionof{figure}{Seven three-factor edits of the open intersection (KITTI 000012). Each group shows the eight combinations of positive intensities.}\label{fig:triples_intersection}\end{minipage}\end{center}

A zero on any axis recovers a view already shown with fewer factors. Each scene therefore contributes one clear image, ten single-factor edits, 36 two-factor edits and 56 three-factor edits across the 21 weather families.

\clearpage
\begin{center}\begin{minipage}{\linewidth}\centering
\includegraphics[width=1\linewidth]{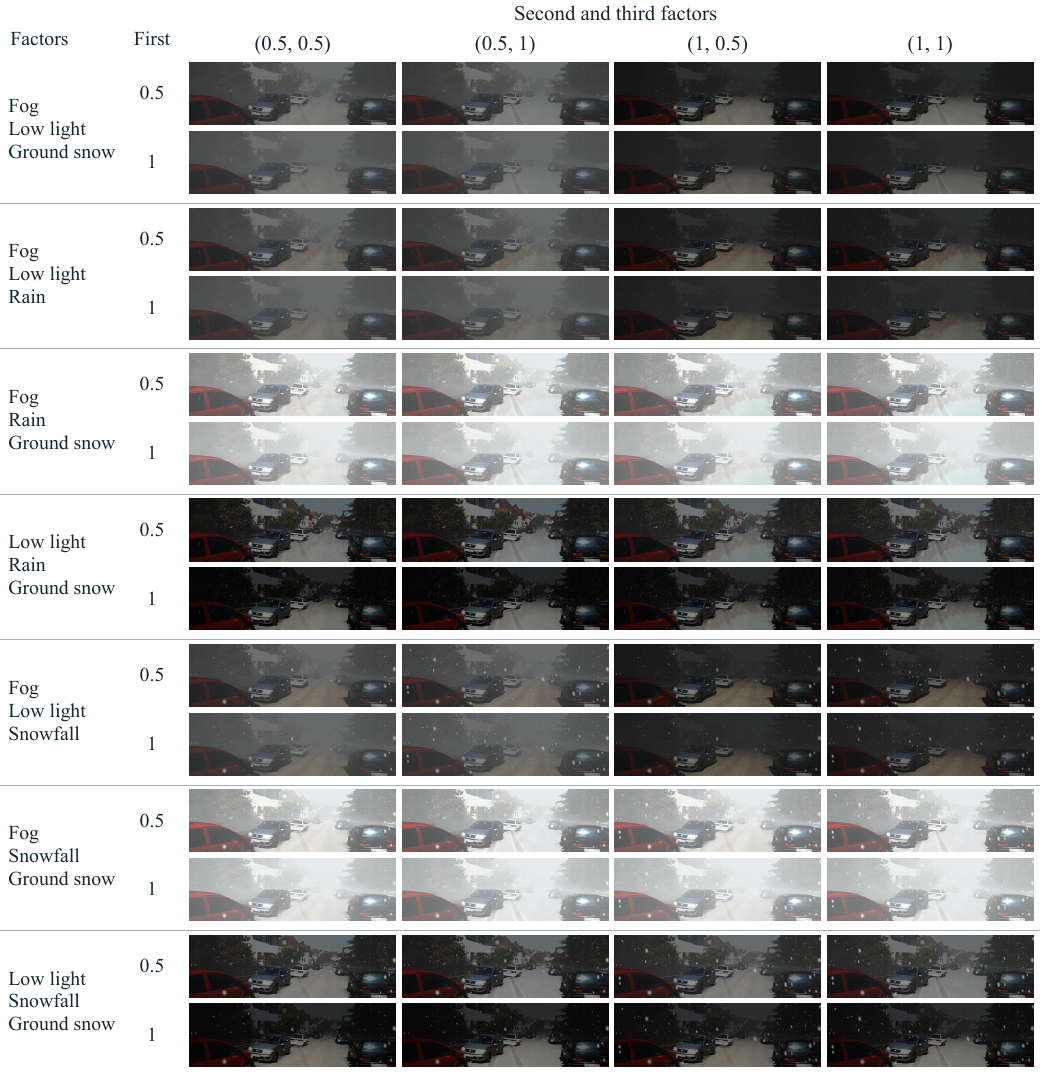}\par
\captionof{figure}{The same seven triples on the residential street (KITTI 000008). Row groups and intensity coordinates match Fig.~\ref{fig:triples_intersection}.}\label{fig:triples_street}\end{minipage}\end{center}

\subsection*{A.2. Choosing the detector performance requirement}

The weather coordinates specify the input. The required F1 determines which settings fall within the detector's capability boundary. At requirements of 0.60, 0.65 and 0.70, respectively, 25, 43 and 43 of the 63 detector--weather combinations include settings on both sides. At 0.65 all three detectors also meet the requirement on the clear cohort. Raising it to 0.70 puts RT-DETR-L below the requirement even on that cohort. At 0.65 the three detectors disagree at 2,626 of 4,201 distinct settings (62.51\%). Comparing YOLO11s and Faster R-CNN, only YOLO11s qualifies at 769 settings and only Faster R-CNN at 28.

\clearpage
\section*{B. Worked examples of weather boundary recovery}
\subsection*{B.1. Selecting queries 8 and 15 on the fog--low-light grid}

Each query measures YOLO11s F1 at one (fog, low light) setting. From $q$ measurements the GP (RBF length 0.24, noise 0.035) estimates the mean $\mu_q(z)$ and residual-scaled standard deviation $s_q(z)$ at each unmeasured node. The probability of meeting $t=0.65$ is $p_z=\Phi(a_z)$, with $a_z=(\mu_q(z)-t)/s_q(z)$. For the two stars in Fig.~\ref{fig:trace}, the seven observations give $a_z=(0.65838-0.65)/0.07774\approx0.108$, hence $p_z=0.543$ at $(0.50,0.875)$. With fourteen observations, $a_z=(0.672973-0.65)/0.024820\approx0.926$ and $p_z=0.823$ at $(0.375,0.875)$.

\noindent\begin{minipage}{\linewidth}\centering
\includegraphics[width=.85\linewidth]{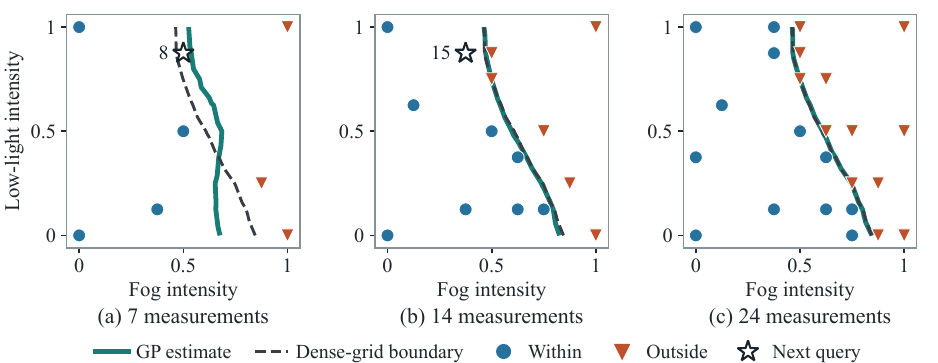}\par
\captionof{figure}{Stars in (a) and (b) mark YOLO11s queries 8 and 15. The solid boundary uses available measurements and the dashed reference uses all 81.}\label{fig:trace}
\end{minipage}\par

The eight smallest $|a_z|$ and eight largest $s_q(z)$ form $C$ after deduplication. The first star ranks sixth and seventh, entering both lists. The second ranks seventh and 35th, entering through $|a_z|$. ALONSO chooses $x^*=\arg\max_{x\in C}\Delta_q(x)$, where the expected reduction in $U=\sum_z p_z(1-p_z)$ is
\[
 \Delta_q(x)=p_x(1-p_x)+\sum_{z\ne x}\big[\Phi_2(a_z,a_z;r_{zx}^2)-p_z^2\big].
\]
The first term resolves $x$ itself. The sum covers every other unmeasured node, using its posterior correlation $r_{zx}$ with the noisy observation at $x$. Four-point quadrature uses the current trend and scale.

Table~\ref{tab:sur_steps} breaks down the gain for queries 8 and 15. For query 8, neighbor $(0.50,0.75)$ has $r_{zx}=0.903$ and contributes 0.148 of the 1.278 gain elsewhere. The runners-up were $(0.625,0.75)$ and $(0.375,1.00)$, respectively.
\begin{center}\begin{minipage}{.73\linewidth}\centering
\captionof{table}{Selection on the fog--low-light grid. Each gain is an expected reduction in $U$.}\label{tab:sur_steps}
\normalsize\renewcommand{\arraystretch}{1.00}\setlength{\tabcolsep}{4pt}
\begin{tabular*}{\linewidth}{@{}l@{\extracolsep{\fill}}rr@{}}\toprule
 & Query 8 & Query 15\\\midrule
Unmeasured nodes & 74 & 67\\
Distinct candidates & 14 & 16\\
Gain at selected node & 0.2482 & 0.1459\\
Gain at other nodes & 1.2781 & 0.2589\\
Total gain & 1.5263 & 0.4048\\
Runner-up total & 1.4987 & 0.3742\\\bottomrule
\end{tabular*}
\end{minipage}\end{center}\smallskip

At query 15 the runner-up offers more gain at its own node (0.1895) but less elsewhere (0.1847). The selected F1 enters the next GP fit. Queries 8 and 15 return 0.64307 (outside) and 0.66807 (within). Linear interpolation on this grid edge gives a crossing at fog 0.465, low light 0.875.

Figure~\ref{fig:paired_queries} compares queries 8--14 after seven shared initial measurements. Both methods fit the same GP. Contour tracing completes the corners of a grid cell around a crossing, so its queries cluster near the lower edge in Fig.~\ref{fig:paired_queries}(a). ALONSO selects the setting with the largest expected reduction in uncertainty across the grid. In Fig.~\ref{fig:paired_queries}(b), its seven queries span several low-light levels.

\clearpage
\raggedbottom
\begingroup\captionsetup{skip=3pt}
\noindent\begin{minipage}{\linewidth}\centering
\includegraphics[width=.58\linewidth]{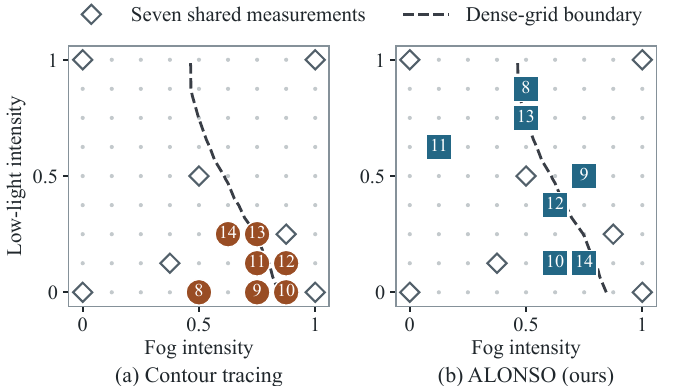}\par
\captionof{figure}{Diamonds mark the shared starting points. Numbers locate queries 8--14 for (a) contour tracing and (b) ALONSO.}\label{fig:paired_queries}
\end{minipage}\par\medskip

\subsection*{B.2. Error trajectories on one- and two-factor grids}

Figures~\ref{fig:queryerrors_one} and~\ref{fig:queryerrors_two} extend the three-factor comparison in main-text \mainThreeFactorFigure{} to one- and two-factor grids.

\noindent\begin{minipage}{\linewidth}\centering
\includegraphics[width=.95\linewidth]{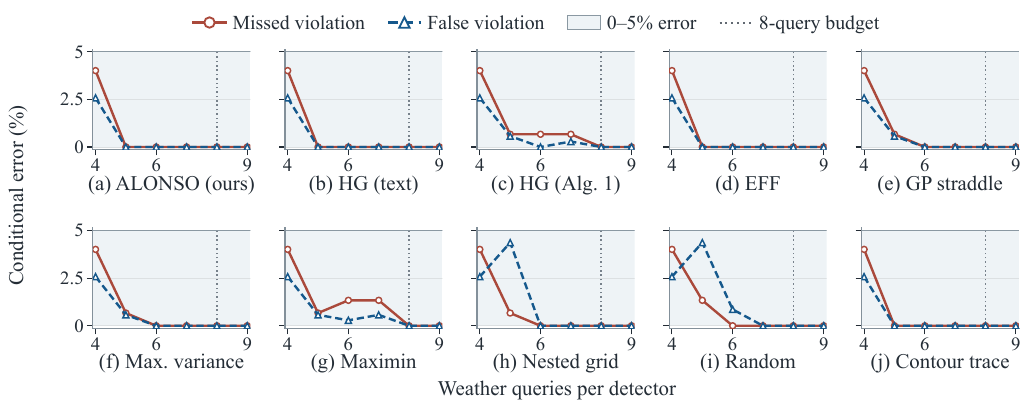}\par
\captionof{figure}{Errors on five one-factor crossing grids. At the eight-query budget, one of nine settings is unmeasured.}\label{fig:queryerrors_one}
\smallskip
\includegraphics[width=.95\linewidth]{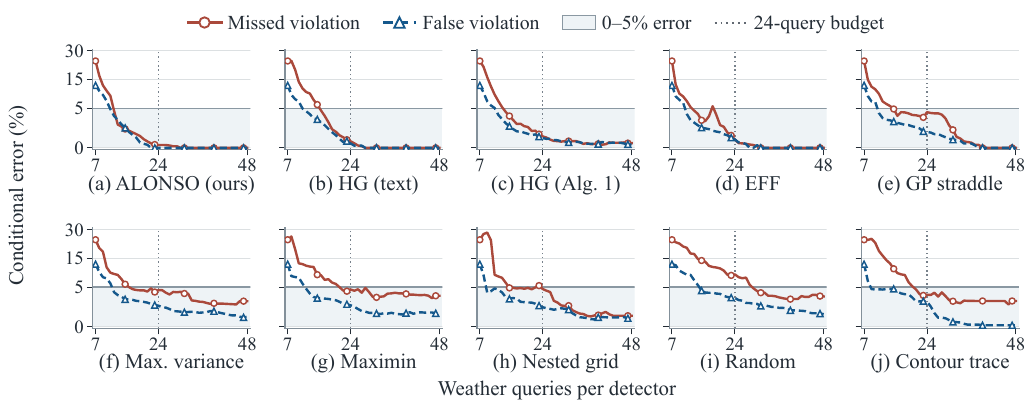}\par
\captionof{figure}{Errors on nineteen two-factor crossing grids. At the 24-query budget, 57 of 81 settings are unmeasured.}\label{fig:queryerrors_two}
\end{minipage}\par\endgroup\smallskip

Circles show the fraction of true failures predicted to pass and triangles the fraction of true passes predicted to fail. Curves average ten initializations within each detector--weather family, giving each crossing family equal weight.

The change in dimension matters for a local search. In one dimension contour tracing reaches the common recovery criterion in all 50 trials. In three dimensions the boundary extends across adjacent cubes. Across 190 crossing trials, 8,719 of its 10,070 adaptive queries between steps 12 and 64 are spent completing the current cell. Bisection uses 530 queries and restarting the search uses 821.

Holding the queried settings fixed separates reconstruction from point selection (Table~\ref{tab:reconstruction}). For each of the 43 crossing detector--weather grids, both reconstructions use the same observations, with budgets of 8, 24 and 64 in one, two and three dimensions. Only unqueried settings enter the error. With the GP fixed and ALONSO choosing the queries, the corresponding error on its own unqueried settings is 0.165\%.

\begin{center}\begin{minipage}{.78\linewidth}\centering
\captionof{table}{Wrong-side decisions at unqueried settings (\%) for two ways of reconstructing the same observations. Lower is better within each row.}\label{tab:reconstruction}
\small\begin{tabular*}{\linewidth}{@{}l@{\extracolsep{\fill}}rr@{}}\toprule
Query selection & Nearest measured label & GP fitted to measured F1\\\midrule
Random & \underline{8.82} & \textbf{3.31}\\
Nested grid & \underline{15.10} & \textbf{2.36}\\\bottomrule
\end{tabular*}\end{minipage}\end{center}

\subsection*{B.3. Online stopping and sustained recovery}

Main-text \mainCostTables{} measure two times on the same query trajectory. The first is the earliest step from which both directional errors stay within 5\% and the fraction of ambiguous nodes stays within 10\% through the query limit. The second is the first time the online rule sees three consecutive updates with at most 10\% ambiguous nodes. A node is ambiguous if it is unmeasured and its 90\% posterior interval spans the requirement.

Table~\ref{tab:stop_example} shows the distinction in one RT-DETR-L run with fog, low light and ground snow. The online rule triggers at query 67. Both observed error counts meet their respective targets there. The next update moves the ambiguous count from 59 to 73, one above the allowable maximum of 72. From query 69 all three conditions hold through query 128. This run therefore contributes 67 queries to the online cost and 69 to the retrospective cost. It counts as passing at the stop and falls outside the sustained-at-stop count.

\begin{center}\begin{minipage}{.89\linewidth}\centering
\captionof{table}{Three successive updates for one RT-DETR-L trial. Error counts are shown against the number of actual failures or passes, and ambiguity against all 729 nodes.}\label{tab:stop_example}
\small\begin{tabular*}{\linewidth}{@{}r@{\extracolsep{\fill}}rrrl@{}}\toprule
Queries & False passes & False alarms & Ambiguous nodes & Event\\\midrule
67 & 6/544 & 3/185 & 59/729 & Online stop\\
68 & 5/544 & 3/185 & 73/729 & Ambiguity exceeds 10\%\\
69 & 5/544 & 2/185 & 70/729 & Sustained recovery begins\\\bottomrule
\end{tabular*}\end{minipage}\end{center}

The relative cost across dimensions also depends on which weather families have a crossing. As a paired check, we match detector and seed along 42 nested chains of one-, two- and three-factor families. Dense grid size divided by mean ALONSO queries is 1.64, 4.44 and 10.62 at retrospective recovery, and 1.20, 4.00 and 10.35 at the online stop. Chains can share runs. Level-set sampling costs depend on the geometry of the response~\cite{bachoc2021}.

\subsection*{B.4. Boundary distance between measured grid settings}

Correctly classifying an off-grid setting and correctly locating its nearest boundary are two distinct outcomes. At fixed budgets on the 304 off-grid development coordinates, ALONSO and HiddenGems (text) disagree on 11 of 9,120 decisions. Independently measured detector F1 agrees with ALONSO on seven and HiddenGems on four. For distance, the reference is the surface interpolated through all measured grid nodes. Table~\ref{tab:geom} gives errors on the same 6,888 predictions with a finite reference and a finite interface under all seven policies in this comparison.

\begin{center}\begin{minipage}{.81\linewidth}\centering
\captionof{table}{Absolute signed-distance errors at off-grid coordinates for three acquisition policies, in normalized weather units.}\label{tab:geom}
\small\begin{tabular*}{\linewidth}{@{}l@{\extracolsep{\fill}}rr@{}}\toprule
Acquisition & Median & 90th percentile\\\midrule
\textbf{ALONSO (ours)} & \textbf{0.00234} & \textbf{0.01221}\\
HiddenGems (text)~\cite{hg} & \underline{0.00260} & \underline{0.01414}\\
GP straddle~\cite{bryan} & 0.00963 & 0.04036\\\bottomrule
\end{tabular*}\end{minipage}\end{center}

Across all 630 ALONSO reconstructions, 200 measured fields meet the requirement everywhere. ALONSO returns this state in all 200 and recovers an interface in each of the other 430. A field with no in-domain interface has a side prediction everywhere and no finite distance to a boundary.

\section*{C. Transferring weather decisions to a frame on a known drive}
\subsection*{C.1. Choosing visual references and updating the transferred field}

ALONSO transfers a weather-response field using a clear target image and the detector identity. The transfer experiment withholds whole frames within each of 40 drives. The target contributes its clear image at zero queries. For each detector and fold, source validation chooses the decoder alone or its combination with the one, three or seven nearest same-drive reference frames, with a reference weight of 0.5, 0.75 or 1. Using the same neighbors for both estimators isolates what fusion adds. Transferring the neighbors' measured responses alone gives 77.47\% accuracy, and adding the decoder improves accuracy by 0.29 percentage points (paired 95\% interval 0.17--0.42).

A local query measures the discrepancy between the transferred F1 and the measured target F1. The four-query schedule measures the clear condition and three joint weather settings on the target frame. Eight queries extend the same schedule. All methods are scored on the same 176 hidden settings across three paired schedules. A query with undefined F1 consumes one measurement and supplies no residual. Source validation excludes all hidden settings and scheduled query positions when choosing the correction.

For the fixed radial basis function (RBF) control, the mean target residual shifts the initial field. An RBF kernel of length 0.24 spreads deviations from that mean according to distance between weather coordinates. ALONSO instead estimates a second-moment matrix of source residuals for each detector. Its diagonal entries estimate the squared error of the visual prior at each weather coordinate. Off-diagonal entries describe which coordinates have errors of similar or opposite sign. Validation compared the unchanged prior, the fixed RBF update, one rank-eight model and two full-rank models. Every fold at every positive budget selected the full matrix with 2\% shrinkage toward its diagonal.

\subsection*{C.2. Visual similarity of reference frames}

Figure~\ref{fig:simthird} shows the accuracy of the image decoder before local queries or retrieval of a reference frame's F1. Within each drive, we rank held-out frames by the distance from their clear-image embedding to the nearest training-frame embedding. Accuracy is 3.97 percentage points higher in the nearest third than in the farthest third (95\% paired drive-bootstrap interval 2.22--5.84). The nearest third has higher accuracy on 32 of 40 drives.

\noindent\begin{minipage}{\linewidth}\centering
\includegraphics[width=.56\linewidth]{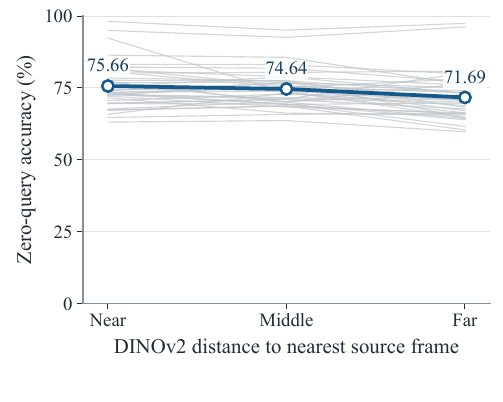}\par
\captionof{figure}{Zero-query decoder accuracy by visual distance to a same-drive training frame. Each drive receives equal weight.}\label{fig:simthird}
\end{minipage}\par\medskip

Reference spacing also affects retrieval. With randomly held-out frames, the nearest visual reference gives 24.12\% wrong-side decisions, compared with 41.43\% for a random same-drive donor. Holding out a contiguous segment gives 38.72\% and 42.53\%, respectively. The drive mean gives 37.80\%. References can precede or follow the segment.

\subsection*{C.3. Weather decisions in consecutive camera views}

All three clear views of KITTI drive 0001 meet the F1 requirement of 0.65 (Fig.~\ref{fig:consecutive_views}). Fog 0.5 alone puts frame 6 outside. At fog 0.25 or low light 0.25 alone, all three qualify. With both at 0.25, YOLO11s detects two, one and two of four eligible vehicles in frames 5, 6 and 7, with no false positives. F1 falls from $2/3$ to $0.40$ and returns to $2/3$.

\noindent\begin{minipage}{\linewidth}\centering
\includegraphics[width=.95\linewidth]{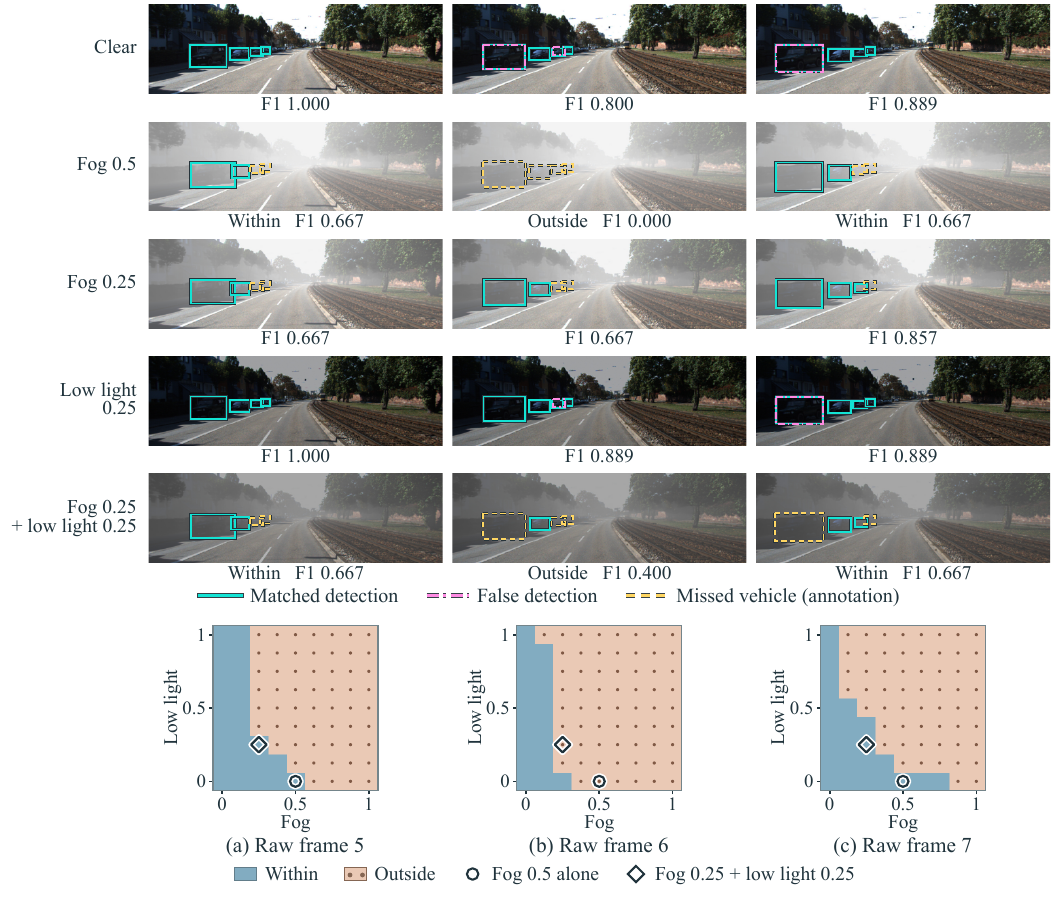}\par
\captionof{figure}{Panels (a)--(c) show scored YOLO11s vehicle detections and missed KITTI annotations above each frame's measured fog--low-light map at zero rain. Circles mark fog 0.5 alone and diamonds mark the joint setting.}\label{fig:consecutive_views}
\end{minipage}\par

In the joint-weather row of Fig.~\ref{fig:consecutive_views}, the nearest parked vehicle is detected in frame 5 and missed in frames 6 and 7. A more distant vehicle is detected in frame 7, bringing F1 back to $2/3$.

\par\smallskip\noindent\makebox[\linewidth][c]{\begin{minipage}{.79\linewidth}\centering\captionsetup{skip=3pt}
\captionof{table}{Settings within the boundary and settings that change side since the preceding frame.}\label{tab:threeframes}
\small\setlength{\tabcolsep}{3pt}\begin{tabular*}{\linewidth}{@{}l@{\extracolsep{\fill}}rrrr@{}}\toprule
 & \multicolumn{2}{c}{2D (81)} & \multicolumn{2}{c}{3D (729)}\\
\cmidrule(lr){2-3}\cmidrule(lr){4-5}
Frame & Within & Changed & Within & Changed\\\midrule
5 & 24 & -- & 77 & --\\
6 & 18 & 6  & 57 & 46\\
7 & 23 & 11 & 132 & 85\\\bottomrule
\end{tabular*}\end{minipage}}\par\smallskip

From frame 5 to 6, six settings leave the fog--low-light qualifying region (Table~\ref{tab:threeframes}). From 6 to 7, eight enter and three leave, nearly restoring its size with a different shape. The three-factor counts also capture changes beyond the zero-rain slice.

\begin{samepage}
Across consecutive raw-frame pairs with both F1 values defined in the eligible drives, the decision changes in 268 of 2,415 clear pairs (11.10\%), 408 of 2,410 at fog 0.5 alone (16.93\%), and 403 of 2,408 at the joint setting (16.74\%). Copying the joint decision from frame 5 to frame 6 would miss the violation.\par
\end{samepage}

\subsection*{C.4. Boundary location and side decisions with a compact decoder}

From a clear image and detector identity, the three heads predict 729 F1 values (A), 729 signed distances (B) or 32 oriented points (C). With a location and unit normal per point and three logits for the field state (all qualifying, all failing or crossing), C uses 195 outputs and 22,729 trainable parameters, compared with 58,507 for A.

Table~\ref{tab:decoders} uses frozen DINOv2 features, five within-drive folds and 32 training epochs. At full training size, B has the lowest wrong-side rate. On the 12,619 fields where all three recover an interface, mean symmetric nearest-interface distance is 0.251 for A, 0.252 for B and 0.161 for C.

In Fig.~\ref{fig:oriented}, 32 points represent 232 measured crossings for YOLO11s on KITTI frame 000069. At low light and rain 0.25, fog 0 qualifies and fog 1 fails. At fog 0, 0.5 and 1, respectively, 61, 38 and 25 of the 81 settings qualify. The reconstructed high-fog band includes five extra settings and misses one. Over the full grid, 32 of 729 decisions change.

\noindent\begin{minipage}{\linewidth}\centering
\includegraphics[width=.91\linewidth]{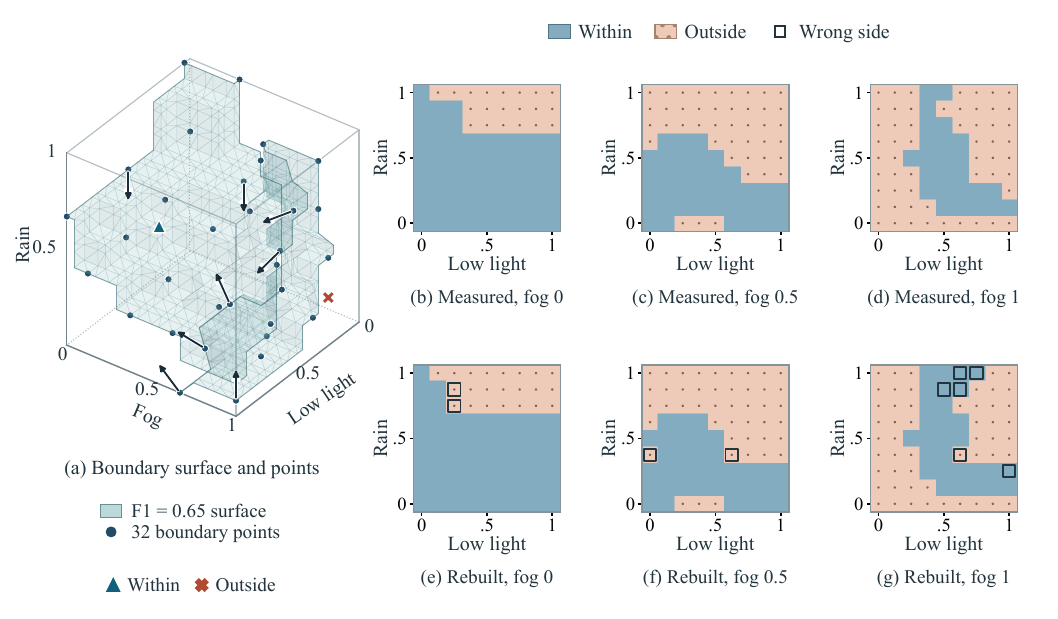}\par
\captionof{figure}{(a) The F1 = 0.65 surface and normals pointing toward the qualifying side. For fog 0, 0.5 and 1, (b--d) show measured decisions and (e--g) the 32-point reconstruction with disagreements marked by squares.}\label{fig:oriented}
\end{minipage}\par\smallskip

Table~\ref{tab:point_count} varies the number of oriented points at full training size. We reconstruct each field from measured crossings or from points predicted from its clear image, scoring the same 13,834,526 nodes with defined F1.

\noindent\begin{minipage}[t]{.455\linewidth}\vspace{0pt}\centering
\captionof{table}{Wrong-side decisions (\%) by fraction of training frames used within each drive.}\label{tab:decoders}
\small\begin{tabular*}{\linewidth}{@{}l@{\extracolsep{\fill}}rrr@{}}\toprule
Decoder & 25\% & 50\% & 100\%\\\midrule
A (F1 field) & \underline{31.19} & \underline{28.79} & \underline{26.70}\\
B (distance field) & \textbf{30.08} & \textbf{27.70} & \textbf{24.97}\\
C (points) & 35.38 & 33.86 & 32.40\\\bottomrule
\end{tabular*}\end{minipage}\hfill
\begin{minipage}[t]{.51\linewidth}\vspace{0pt}\centering
\captionof{table}{Wrong-side decisions (\%) by point count and source. Lower is better within each row.}\label{tab:point_count}
\small\begin{tabular*}{\linewidth}{@{}l@{\extracolsep{\fill}}rrrr@{}}\toprule
Point source & 16 & 32 & 64 & 128\\\midrule
Measured & 12.33 & 8.35 & \underline{4.70} & \textbf{1.81}\\
Image & 33.05 & \underline{32.40} & \textbf{32.10} & \textbf{32.10}\\\bottomrule
\end{tabular*}\end{minipage}\par\medskip

Using more measured crossings improves reconstruction at every point count tested. Across 40 equally weighted drives, increasing the learned point count from 16 to 32 lowers error by 0.63 percentage points (95\% paired interval 0.43--0.83). The 64- and 128-point heads both give 32.10\% pooled error at the reported precision. With 32 points, adding supervision on the qualifying side at every defined training node reduces the learned decoder's wrong-side rate from 32.40\% to 29.34\%.

\end{document}